\documentclass{article}

\usepackage{microtype}
\usepackage{graphicx}
\usepackage{subcaption}
\usepackage{booktabs} % for professional tables
\usepackage{hyperref}
\usepackage{placeins}

\usepackage[accepted]{icml2026}

\usepackage{amsmath}
\usepackage{amssymb}
\usepackage{mathtools}
\usepackage{amsthm}

\usepackage[capitalize,noabbrev]{cleveref}

\theoremstyle{plain}

\theoremstyle{definition}

\theoremstyle{remark}

\usepackage[textsize=tiny]{todonotes}

\icmltitlerunning{Mechanistic origins of catastrophic forgetting: why RL preserves circuits better than SFT?}

\begin{document}

\twocolumn[
  \icmltitle{Mechanistic origins of catastrophic forgetting: \\
    why RL preserves circuits better than SFT? }



  \icmlsetsymbol{equal}{*}

  \begin{icmlauthorlist}
    \icmlauthor{Jeanmely Rojas Nunez}{alg}
    \icmlauthor{Viraj Sawant}{alg}
    \icmlauthor{Nathan Allen}{alg}
    \icmlauthor{Nomgondalai Amgalanbaatar
}{alg}
    \icmlauthor{Yannis Zongo}{alg}
    \icmlauthor{Vasu Sharma}{alg}
    \icmlauthor{Maheep Chaudhary}{indep}
  \end{icmlauthorlist}

  \icmlaffiliation{alg}{Algoverse AI Research}
  \icmlaffiliation{indep}{Independent}
  \icmlcorrespondingauthor{Jeanmely Rojas Nunez}{jeanmelyrojasnunez@gmail.com}
  
  \icmlkeywords{Machine Learning, ICML}

  \vskip 0.3in
]



 % no special notice (required even if empty)
\printAffiliationsAndNotice{\icmlEqualContribution}

\begin{abstract}
  Fine-tuning large language models (LLMs) frequently induces catastrophic forgetting of prior capabilities. Recent work has shown that reinforcement learning (RL) can retain prior capabilities more effectively than supervised fine-tuning (SFT), and this effect is often interpreted through policy proximity to the base policy \cite{shenfeld2025rl}. We revisit this behavioral account at the mechanistic level and ask whether RL's advantage is associated with stronger preservation of internal computational circuits, even when output-space KL alone does not fully predict forgetting. We introduce differential circuit vulnerability, a head-level measure of how much a circuit degrades under fine-tuning, and use it to compare RL and SFT on Qwen2.5-3B-Instruct adapted to scientific question-answering. We find a mechanistic trade-off in this setting: SFT adapts more rapidly to the target task but produces greater circuit disruption and forgetting of prior capabilities, whereas RL preserves a larger fraction of the base circuit at the cost of slower task adaptation. These findings suggest that circuit preservation may help explain RL's robustness to catastrophic forgetting in our experiments. We released our code here: \url{https://github.com/rl-sft-circuit-research/differential-circuit-vulnerability}.
\end{abstract}

% Teaser figure: high-level conceptual overview placed right after the abstract.
% Spans both columns via figure*. Quantitative results appear later in Section 4.

\section{Introduction}
Adapting large language models (LLMs) to new downstream tasks often leads to catastrophic forgetting: gains on a new objective come at the expense of prior capabilities. As models are increasingly expected to update continuously and adapt to new domains, mitigating such degradation has become a central challenge in post-training. Prior work establishes that the choice of adaptation objective strongly shapes this trade-off: RL often preserves 
prior capabilities more effectively than SFT \cite{shenfeld2025rl,hu2025open}. This advantage is commonly interpreted through policy proximity to the pretrained model, often measured by KL divergence. However, output-space proximity may not fully explain retention in all settings. We therefore revisit this account mechanistically and ask whether retention is better aligned with preservation of internal circuits than with output-space KL alone.

% \begin{figure}[H]
%     \centering
%     \includegraphics[width=0.92\columnwidth]{icml2026/Trajectory_Plot_ContrastZ.pdf}
%     \caption{\textbf{Conceptual overview.} SFT and RL models diverge from the base model circuit along different trajectories in the preservation--performance plane. RL drifts conservatively and retains more of the base circuit; SFT specializes aggressively, achieving higher task scores at the cost of greater circuit disruption. We quantify this trade-off in Section~\ref{sec:results}.}
%     \label{fig:trajectory}
% \end{figure}

\begin{figure}[H]
    \centering
    \includegraphics[width=0.97\columnwidth]{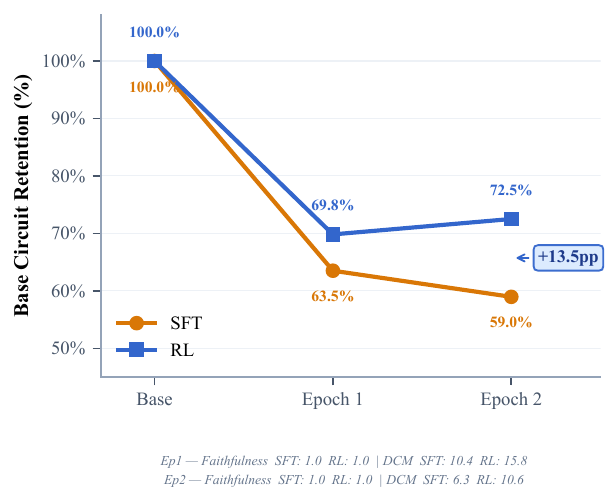}
    \caption{\textbf{Circuit retention trajectories during high-NTS training.}
Starting from $100\%$ base-circuit retention, SFT (orange) and RL (blue) diverge over two training epochs.
SFT drops to $63.5\%$ after Epoch~1 and further to $59.0\%$ by Epoch~2.
RL drops to $69.8\%$ after Epoch~1 and increases to $72.5\%$ by Epoch~2, a $13.5$ percentage-point difference at Epoch~2.
Footer values report per-epoch circuit faithfulness, both $\approx 1.0$, and Differential Causal Mediation (DCM).
RL yields higher DCM at both epochs: $15.8$ vs.\ $10.4$ at Epoch~1 and $10.6$ vs.\ $6.3$ at Epoch~2, indicating that the retained circuit components exert greater causal influence on model outputs under RL than under SFT.}
    \label{fig:circuit-trajectory}
\end{figure}

% \begin{figure}[H]
%     \centering
%     \includegraphics[width=0.97\columnwidth]{icml2026/Fig2_Trajectory.pdf}
%     \caption{\textbf{Temporal overview of task acquisition.} SFT and RL models learn new task capabilities at very different speeds during training.  SFT specializes aggressively, rapidly achieving higher task scores within a small number of epochs. On the other hand, RL (GRPO) suffers from conservative drift and gradually improves its performance on the new task over more iterations. We quantify this trade-off in Section~\ref{sec:results}.}
%     \label{fig:trajectory}
% \end{figure}

In parallel, mechanistic interpretability has shown that model capabilities are implemented by internal computational \textit{circuits} composed of attention heads, MLP layers, and residual pathways. Fine-tuning may therefore succeed or fail depending on whether it preserves these structures or disrupts them \cite{prakash2024finetuning}. This motivates our central question: \textbf{is RL's retention advantage associated with stronger preservation of task-relevant circuits relative to SFT?}

We examine post-training through the lens of circuit preservation. We introduce \textit{differential circuit vulnerability}, the relative susceptibility of internal computational subgraphs to degradation under different training objectives. Using this framework, we compare RL and SFT on matched adaptation tasks and identify which circuits are preserved or disrupted. Figure~\ref{fig:circuit-trajectory} previews the central finding in our setting: over the course of training the two objectives trace diverging circuit-retention trajectories, with RL preserving a larger fraction of the base circuit. We characterize this trade-off mechanistically in the sections that follow.

We instantiate our analysis on Qwen2.5-3B-Instruct using a two-stage protocol: adaptation on scientific question answering, followed by retention evaluation across a broad benchmark suite spanning commonsense reasoning, factuality, and instruction following. The results expose a trade-off in this setting: SFT adapts faster to the new objective but at the cost of greater circuit disruption, whereas RL preserves more of the base circuit and of prior capabilities, occasionally at the expense of under-optimizing the new task.

These results suggest that continual adaptation may be usefully understood not merely as parameter optimization, but as the \textit{selective preservation and modification} of circuits encoding skills. This perspective points toward a possible mechanistic account of why RL appears more conservative than SFT during post-training.

\section{Related Work}

\paragraph{Reinforcement Learning, Supervised Fine-Tuning, and Forgetting.}
Supervised fine-tuning (SFT) and reinforcement learning (RL) are the dominant post-training methods for adapting foundation models, yet their effects on previously acquired capabilities remain incompletely understood. SFT optimizes imitation on a target distribution, whereas RL updates behavior through reward optimization and often produces qualitatively different parameter changes. Recent studies suggest that, under comparable settings, RL can preserve pretrained capabilities better than SFT, indicating that the training objective influences both adaptation and forgetting \cite{shenfeld2025rl,hu2025open}.

The RL method used in this work, Dr.GRPO, builds on Group Relative Policy Optimization \cite{shao2024deepseekmath}, a policy gradient algorithm that estimates advantages by normalizing rewards within a sampled group of completions rather than relying on a learned value function. Unlike PPO-style objectives, GRPO omits an explicit KL penalty, instead relying on group-relative normalization and conservative learning rates to limit policy drift. Dr.GRPO extends this framework to domain-specific settings by replacing a single generic reward with task-adapted correctness signals. This design allows the reward signal to remain tightly coupled to the structure of the target domain without requiring a separately trained reward model.

\paragraph{Mechanistic Interpretability and Circuit Analysis.}
Mechanistic interpretability explains model behavior through concrete components such as attention heads, MLP layers, and circuits. Intervention-based methods, including activation patching, path patching, and masking, have shown that many behaviors can be localized to sparse causal subnetworks, and recent work provides a unifying causal-abstraction framework for these techniques \cite{geiger2025causal}. Related work further suggests that fine-tuning frequently modifies existing mechanisms rather than replacing them entirely \cite{davies2023discovering,prakash2024finetuning}, that training procedures themselves can be shaped to yield more modular circuit structure \cite{golechha2025modular}, and that some circuits are alignment-critical and should be preferentially preserved during model modification \cite{patel2025alignment}.

\paragraph{Evaluating Retention Across Capabilities.}
Retention cannot be assessed on the adaptation task alone. Models may improve on a new objective while degrading reasoning, factuality, instruction following, or code generation. For this reason, prior work increasingly relies on diverse benchmark suites. Common evaluations include MMLU, HellaSwag, and WinoGrande for reasoning and commonsense inference, and TruthfulQA, IFEval, and HumanEval for factuality, instruction following, and coding \cite{hendrycks2021measuring,zellers2019hellaswag,lin2022truthfulqa}.

\paragraph{Research Gap and Positioning.}
Despite progress in both post-training and interpretability, the two literatures remain loosely connected. Optimization studies typically report benchmark outcomes without explaining the internal causes of forgetting, while interpretability studies rarely compare learning objectives. We address this gap by comparing RL and SFT through the lens of circuit preservation, asking which objective better maintains the causal structures underlying pretrained capabilities.
% Note use of \abovespace and \belowspace to get reasonable spacing
% above and below tabular lines.
\section{Methodology}

We test whether RL's retention advantage is associated with stronger preservation of task-relevant internal circuits than SFT. Our pipeline comprises three phases: (I) compare retention, output-space drift, and circuit preservation between SFT and RL; (II) identify circuits in each model; and (III) compare how post-training reshapes these circuits.

We begin from a pretrained model $\pi_{\mathrm{base}}$. We first train an SFT model $\pi_{\mathrm{SFT}}$ via completion-only supervision, then refine it with Dr.GRPO to obtain $\pi_{\mathrm{RL}}$. Thus, our comparison isolates the effect of continuing post-training with RL beyond SFT $\pi_{\theta}\in\{\pi_{\mathrm{SFT}},\pi_{\mathrm{RL}}\}$.

To quantify behavioral drift from the base model, we compute the expected KL divergence on retention tasks, where lower values correspond to less distributional shift.

\begin{equation}
\mathbb{E}_{x\sim\tau}
\left[
D_{\mathrm{KL}}
\left(
\pi_{\mathrm{base}}(\cdot|x)\,\middle\|\,\pi_{\theta}(\cdot|x)
\right)
\right].
\label{eq:kl-drift}
\end{equation}

\subsection{Phase I: Reproducing Distributional Shift Effects}

We first compare retention, output-space KL, and circuit preservation across SFT and RL in order to test whether behavioral retention is better aligned with policy proximity or with internal circuit preservation \cite{shenfeld2025rl}. Models are trained on a downstream Task A and evaluated on a separate suite of retention benchmarks (Task B).

\paragraph{SFT.}
We fine-tune $\pi_{\mathrm{base}}$ with a completion-only cross-entropy loss.

\paragraph{RL.}
We refine $\pi_{\mathrm{SFT}}$ with Dr.GRPO. The model samples candidate completions, receives binary rewards, computes normalized group-relative advantages, and updates the policy through a weighted log-probability objective. We use a group size of $64$, two refinement steps ($\mu=2$), and no explicit KL penalty.

\subsection{Phase II: Circuit Identification via Differential Binary Masking}

We analyze circuits at the attention-head level using Differential Binary Masking (DBM) \cite{chaudhary2024evaluating}. DBM learns a mask over heads that interpolates between base and counterfactual activations: $\tilde{a}_h=(1-m_h)a_h^{\mathrm{base}}+m_h a_h^{\mathrm{source}}$

where $m_h\in[0,1]$. Annealing pushes the masks toward binary selections, yielding sparse causal circuits.

\paragraph{Triplets.}
For chemistry QA, we construct triplets $(x_{\mathrm{base}},x_{\mathrm{source}},y_{\mathrm{target}})$ corresponding to three counterfactual hypotheses: answer-key swaps, molecule swaps, and task-type swaps.

\paragraph{Objective.}
Masks are optimized to increase the probability of the target answer while remaining sparse: $\mathcal{L}_{\mathrm{DBM}}
=
-\log P(y_{\mathrm{target}}|x,\tilde{a})
+\lambda\sum_h m_h$

\paragraph{Scoring.}
We score answers using the geometric mean of token probabilities 
\begin{equation}
p(y|x)=
\exp\left(
\frac{1}{T}\sum_{i=1}^{T}\log P(y_i|x,y_{<i})
\right).
\label{eq:geom-prob}
\end{equation}

Circuit discovery is run independently for $\pi_{\mathrm{base}}$, $\pi_{\mathrm{SFT}}$, and $\pi_{\mathrm{RL}}$.

\subsection{Phase III: Cross-Model Circuit Comparison}

We assess how strongly the discovered circuits remain functional after post-training. Circuit faithfulness is defined as

\begin{equation}
\mathrm{Faithfulness}(\mathcal{C},M)=\frac{F(\mathcal{C}|M)}{F(M)}.
\label{eq:faithfulness}
\end{equation} 

with values close to $1$ indicating that the circuit recovers most of the model's behavior.

For each head $h$, we compare DBM mask values against those of the base model:

\begin{equation}
 \Delta m_h(M)=m_h^M-m_h^{\mathrm{base}},
\label{eq:maskshift}
\end{equation} 

where $M\in\{\pi_{\mathrm{SFT}},\pi_{\mathrm{RL}}\}$. This identifies heads that are preserved, amplified, or weakened by training.

We define \textit{vulnerable} heads as those more degraded under SFT than under RL:  $\mathcal{C}_{\mathrm{vuln}}
=
\{h:\; m_h^{\mathrm{SFT}}<m_h^{\mathrm{RL}}-\delta\}.$
 
\section{Experiments}

Our experiments evaluate whether RL's retention advantage is associated with stronger preservation of task-relevant circuits than SFT. We address two questions: (I) How do forgetting and behavioral drift compare between SFT and RL; and (II) are retention differences reflected in stronger circuit faithfulness, more stable head-level contributions, and more distributed circuits?

\subsection{Experimental Setup}
All experiments use Qwen2.5-3B-Instruct. The fine-tuning task (Task A) is scientific question answering. Retention (Task B) is measured on benchmarks spanning commonsense reasoning, factuality, instruction following, and code generation, since forgetting can be capability-specific. We compare three systems: the pretrained base model, standard SFT, and RL with Dr.GRPO. Behavior is measured with downstream accuracy and KL divergence from the base model. The mechanistic analysis uses circuit faithfulness, head-level mask shifts $\Delta m_h = m_h^M - m_h^{\text{base}}$ derived from cross-model DBM mask comparison, and necessity/sufficiency interventions; necessity measures the log-probability drop when a head is disrupted, while sufficiency measures how much behavior is recovered when only that head is kept. Together, these metrics distinguish distributed contributors from critical bottlenecks.
\paragraph{New Task Score (NTS).}
We define the \textit{New Task Score} (NTS) as the model's accuracy on Task~A
(SciKnowEval science QA), computed as the fraction of questions for which
the model assigns the highest geometric-mean token probability
(Eq.~\ref{eq:geom-prob}) to the correct answer choice. Table~\ref{tab:experimental_setup} summarizes the full setup.

\subsection{Results}
\label{sec:results}
DBM identifies a base circuit comprising 297 attention heads ($51.6\%$ of all attention heads). After adaptation, the SFT model exhibits structural compression to roughly 265 heads ($46.0\%$), whereas the RL model retains approximately 296 heads ($51.4\%$), close to the base model's 297. The same disparity is reflected in base-circuit overlap: the RL model preserves about $68\%$ of base heads, substantially more than the SFT model's $52\%$ (Figure~\ref{fig:all-graphs}).

% \begin{figure}[H]
%      \centering
%      \begin{subfigure}[b]{0.9\columnwidth}
%          \includegraphics[width=\textwidth, trim={0 2cm 0 2cm}, clip]{icml2026/image (2).pdf}
%          \caption{Circuit Count}
%      \end{subfigure}
%      \hfill
%      \begin{subfigure}[b]{0.9\columnwidth}
%          \includegraphics[width=\textwidth, trim={0 7cm 0 7cm}, clip]{icml2026/image (3).pdf}
%          \caption{Circuit Overlap}
%      \end{subfigure}
     
%      \caption{Figure 1(a): Circuit Size Comparison — The Base model uses a functional circuit of 297 components, which the SFT process prunes down to a 265-component structure, while the RL model uses a denser, distributed sub-network of 296 components.Figure 1(b): Base Circuit Preservation — The RL model preserves 67.34\% of the original architecture, compared to the SFT model's 51.51\% overlap. This 15.83\% greater preservation rate describes why RL is more immune to catastrophic amnesia.}
% \end{figure}

Figure~\ref{fig:circuit-trajectory} traces this divergence over the course of the training runs that yield the high new-task score models. Both objectives begin from full base-circuit retention, but their trajectories separate immediately: after the first epoch SFT retains only $63.5\%$ of base heads compared to RL's $69.8\%$, and by the second epoch the gap widens to $13.5$ percentage points ($59.0\%$ vs.\ $72.5\%$). Notably, RL's retention \emph{increases} between epochs, suggesting that continued RL training can recover previously disrupted circuit components rather than monotonically degrading them. Throughout training, RL also maintains consistently higher DCM scores ($15.8$ vs.\ $10.4$ at Epoch~1; $10.6$ vs.\ $6.3$ at Epoch~2), indicating that RL's preserved circuit carries substantially more causal signal on the answer-key counterfactual. The two objectives thus diverge along different axes of the same trade-off: SFT collapses adaptation into a small specialist set that discards much of the base circuit, while RL spreads adaptation across a broader, more causally engaged subgraph that retains substantially more of the original computation.

Circuit faithfulness exceeds $1.0$ for all three models ($1.02$ base, $1.04$ SFT, $1.12$ RL), confirming that the extracted subgraphs recover the full model's behavior on the target task.

 \begin{figure}[t]
    \centering
     \includegraphics[width=\columnwidth]{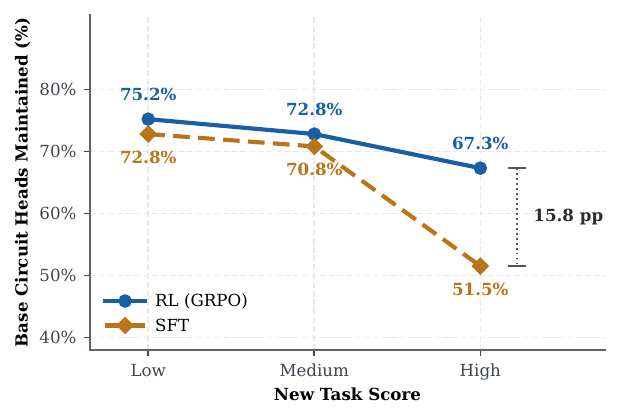}
    \caption{\textbf{Performance--preservation trade-off across NT levels.} SFT (dashed) shows a steeper decline in circuit retention as new-task score increases, while RL (solid) declines more gradually. At peak new-task performance, RL retains $15.8$ percentage points more of the base circuit than SFT. Low ($<$50\%), Medium (50--65\%), High ($>$65\%).} 
     \label{fig:nts-tradeoff}
 \end{figure}

\begin{figure}[t]
    \centering
    \includegraphics[width=0.90\linewidth, trim={0 0cm 0 0cm}]{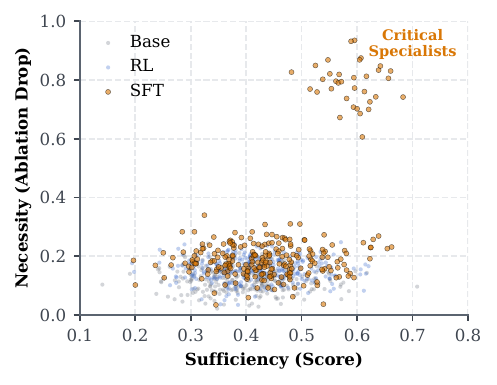}
    \caption{\textbf{Head Role Distribution Under Base, SFT, and RL Training.} SFT produces a cluster of heads with high necessity and high sufficiency scores, corresponding to the upper-right region, relative to the base model. RL yields a distribution that more closely overlaps with the base model, with fewer heads concentrated in the high-necessity, high-sufficiency region.}

    \label{fig:critics}
\end{figure}

\subsection{Functional Importance and Circuit Drift}

Beyond aggregate circuit size, we ask whether fine-tuning preferentially modifies the heads that matter most to the base model. Figure~\ref{fig:critics}
shows the necessity--sufficiency landscape: SFT produces a tight cluster of
``critical specialists''---heads with both high necessity and high
sufficiency---while RL maintains a flatter distribution that closely tracks the base model. This is the mechanistic signature of compression: SFT consolidates new-task behavior into a small set of high-importance heads,
whereas RL spreads adaptation across many components without any single one
becoming indispensable.

Figure~\ref{fig:necessity-delta} probes this directly, plotting per-head necessity against the mask shift $\Delta m_h$ (Eq.~\ref{eq:maskshift}). If
either objective preferentially edited the most important heads, we would
expect a positive correlation. We do not observe one: SFT shows a weak
negative trend ($r{=}{-}0.125$, $p{=}0.12$) and RL is essentially flat ($r{=}0.022$). Read together with Figure~\ref{fig:critics}, the picture is consistent across both models: neither objective rewires the circuit by tracking base-model necessity. SFT instead reorganizes broadly while elevating a new set of specialists, whereas RL adjusts uniformly across heads without producing any such concentration.

\subsection{Discussion}

Taken together, our findings suggest that SFT exhibits a mechanistic ``breaking point'' in this setting. SFT and RL are comparable at low levels of adaptation, suggesting that both objectives can initially improve new-task performance without substantially disrupting the base circuit. In the high-NTS regime, however, SFT shifts from conservative adaptation to aggressive circuit reconfiguration, while RL remains comparatively stable. This pattern is consistent with RL preserving and reusing existing representations more effectively than SFT, which appears to overwrite portions of the base circuit under stronger new-task pressure. The epoch-level trajectory in Figure~\ref{fig:circuit-trajectory} reinforces this interpretation: SFT's retention declines monotonically across epochs, whereas RL's retention recovers after an initial drop, suggesting that RL may consolidate and restore circuit structure during continued training. Confirming this interpretation will require experiments across additional models, tasks, and optimization settings.

We also observe that the RL model maintains greater prior-task retention and circuit preservation despite having a larger output-space Kullback--Leibler (KL) divergence than the SFT model. This suggests that output-space metrics may not reliably predict internal forgetting in this setting, consistent with broader observations that surface behavior and internal representational transfer can diverge during post-training procedures such as distillation \cite{konigquantifying}. In particular, RL may alter the model's output distribution while preserving much of the underlying computation responsible for prior-task performance. By contrast, SFT can appear less divergent at the output level while still reorganizing the internal circuits more aggressively.

\begin{figure}[t]
    \centering
    \includegraphics[width=\columnwidth]{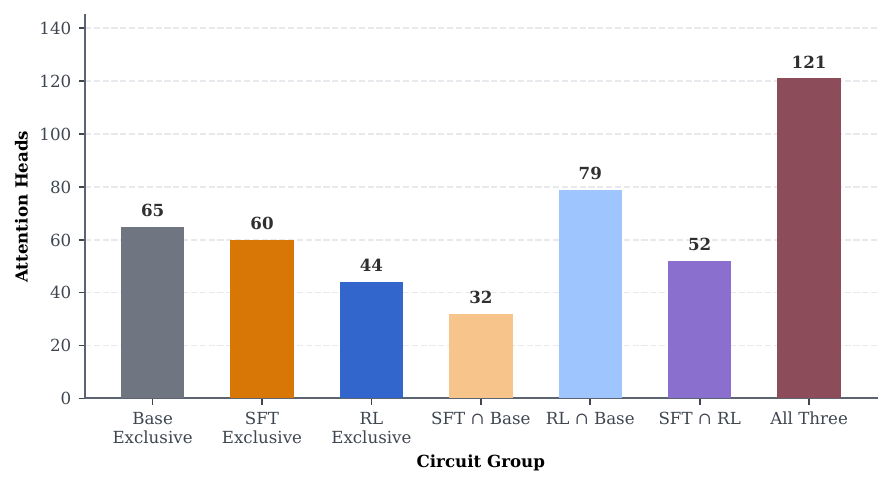} 
    \caption{\textbf{Attention Head Overlap Between Base, SFT, and RL.} Each bar segment reports the number of attention heads belonging exclusively to one model, shared by exactly two models, or present in all three. RL shares more heads with the base model than SFT does, while SFT retains the fewest heads in common with either the base model or RL.}
    \label{fig:all-graphs} 
\end{figure}

The gap between the SFT and RL circuit sizes ($\sim265$ vs.\ $\sim295$ heads), depicted in Figure~\ref{fig:all-graphs}, supports a characterization of SFT as a ``compressor'' and RL as a ``distributed adaptor.'' SFT concentrates new-task behaviour into a smaller, more specialized circuit, while RL preserves a broader set of base-model heads during adaptation.

% \vspace{-20pt}

% \FloatBarrier
\section{Conclusion}
We asked whether RL's retention advantage over SFT is associated with stronger preservation of internal circuits. In our Qwen2.5-3B-Instruct science-QA setting, our behavioral and mechanistic analyses are consistent with this hypothesis: RL preserves more of the original circuit and maintains stronger functional faithfulness, even when output-space KL does not indicate closer proximity to the pretrained model. By contrast, SFT adapts faster but shows greater circuit reorganization and more forgetting in our experiments.

These results point to a potential trade-off in post-training: rapid specialization versus stable reuse of existing mechanisms. In our setup, SFT optimizes the new objective efficiently but can disrupt circuits supporting prior skills, while RL appears more conservative, preserving those circuits at the cost of smaller task gains. More broadly, future adaptation methods may benefit from combining efficient learning with selective circuit preservation, toward models that continue improving without forgetting.

\section{Limitations and Future Work}
This study examines only one model (Qwen2.5-3B-Instruct), limiting the generalizability of our findings. Future work should validate these trade-offs across diverse architectures: Gemma, Mistral, Llama, and Pythia, and at varying parameter scales.
Our circuit analysis is further constrained to attention heads and a narrow task set. Expanding to MLP layers, residual-stream features, and broader capability domains: multilingual reasoning, factual recall, safety, and tool use, would yield a more complete mechanistic picture.

\nocite{langley00}

\bibliography{main}
\bibliographystyle{icml2026}

%%%%%%%%%%%%%%%%%%%%%%%%%%%%%%%%%%%%%%%%%%%%%%%%%%%%%%%%%%%%%%%%%%%%%%%%%%%%%%%
%%%%%%%%%%%%%%%%%%%%%%%%%%%%%%%%%%%%%%%%%%%%%%%%%%%%%%%%%%%%%%%%%%%%%%%%%%%%%%%
% APPENDIX
%%%%%%%%%%%%%%%%%%%%%%%%%%%%%%%%%%%%%%%%%%%%%%%%%%%%%%%%%%%%%%%%%%%%%%%%%%%%%%%
%%%%%%%%%%%%%%%%%%%%%%%%%%%%%%%%%%%%%%%%%%%%%%%%%%%%%%%%%%%%%%%%%%%%%%%%%%%%%%%

\newpage
\appendix
\onecolumn
\section{Appendix}
\begin{table}[!h]
\centering
\caption{Experimental Setup}
\label{tab:experimental_setup}
\begin{tabular}{ll}
\toprule
\textbf{Component} & \textbf{Details} \\
\midrule
\textbf{Model} & Qwen2.5-3B-Instruct \\
\midrule
\textbf{Task A (Fine-tuning)} 
& SciKnowEval~\cite{feng2025sciknoweval} (Science Q\&A) \\
 
\midrule
\textbf{Task B (Retention)} 
& HellaSwag~\cite{zellers2019hellaswag}, TruthfulQA~\cite{lin2022truthfulqa}, \\
& MMLU~\cite{hendrycks2021measuring}, IFEval~\cite{zhou2023instruction}, \\
& WinoGrande~\cite{sakaguchi2019winogrande}, HumanEval~\cite{chen2021evaluating} \\
\midrule
\textbf{Metrics} 
& KL divergence (Eq.~\ref{eq:kl-drift}) \\
& Adaptation performance and retention accuracy \\
& Circuit faithfulness (Eq.~\ref{eq:faithfulness}) \\
& Head-level mask shift $\Delta m_h(M)$ from CMAP (Eq.~\ref{eq:maskshift}) \\
\midrule
\textbf{Baselines} 
& Base model (no fine-tuning) \\
& SFT (standard) \\
& RL (Dr.GRPO) \\
\bottomrule
\end{tabular}
\end{table}

\begin{figure}[htbp]
    \centering
    \includegraphics[width=0.8\textwidth]{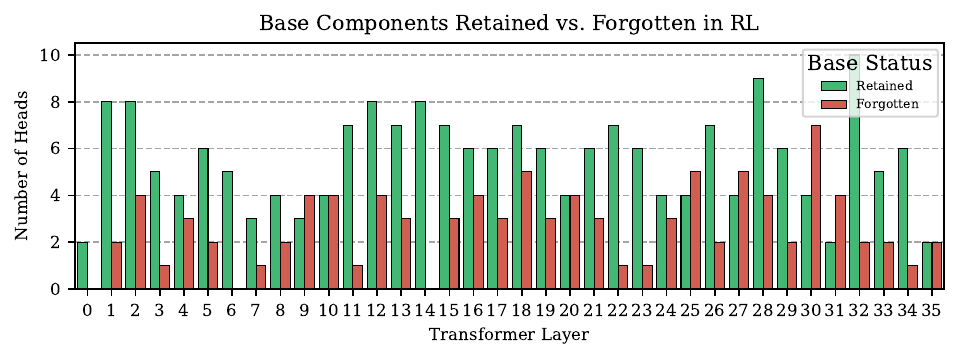}
    \caption{\textbf{Layer-wise Circuit Retention Under RL Training.} Each bar reports the number of attention heads per layer categorized as retained or forgotten relative to the base model circuit, across all 36 transformer layers. Retained heads outnumber forgotten heads in the majority of layers.}
    
    \label{fig:rl_plot}
\end{figure}

\begin{figure}[htbp]
    \centering
    \includegraphics[width=0.8\textwidth]{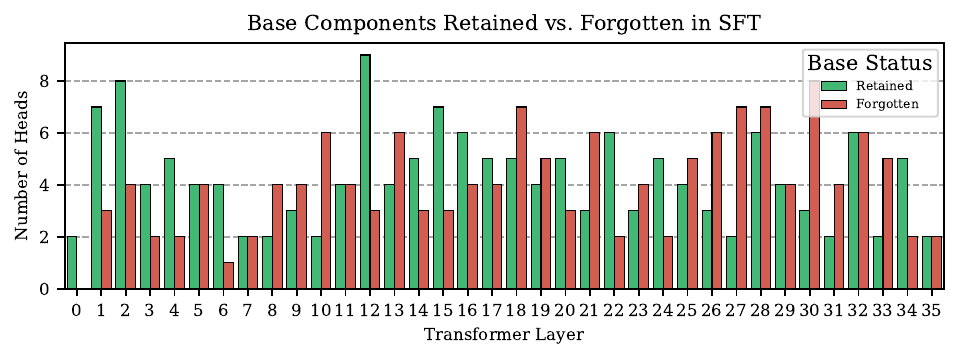}
    \caption{\textbf{Layer-wise Circuit Retention Under SFT Training.} Each bar reports the number of attention heads per layer categorized as retained or forgotten relative to the base model circuit, across all 36 transformer layers. Forgotten heads appear across all layers, with higher counts in the mid-to-late layers.}
    
    \label{fig:sft_plot1}
\end{figure}

\begin{figure}[H]
    \centering
    \includegraphics[width=\columnwidth]{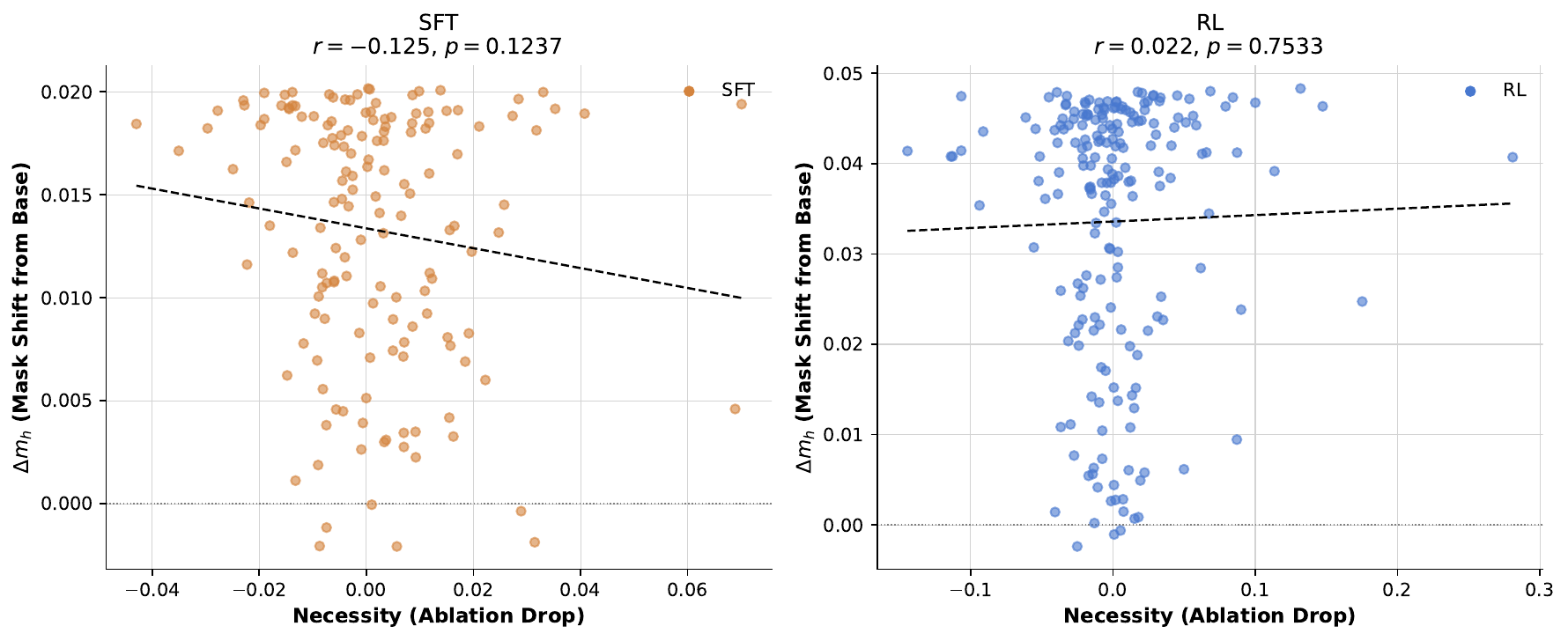}
    \caption{\textbf{Per-Head Necessity vs.\ $\Delta m_h$: SFT vs.\ RL.} Each point represents one attention head, plotting necessity score against $\Delta m_h$ as defined in Eq.~\ref{eq:maskshift}. Neither model shows a positive correlation between necessity and mask shift. SFT exhibits a weak negative correlation, $r = -0.125$, and RL shows a near-zero correlation, $r = 0.022$.}
    \label{fig:necessity-delta}
\end{figure}

\begin{figure}[htbp]
    \centering
    \includegraphics[width=0.6\textwidth]{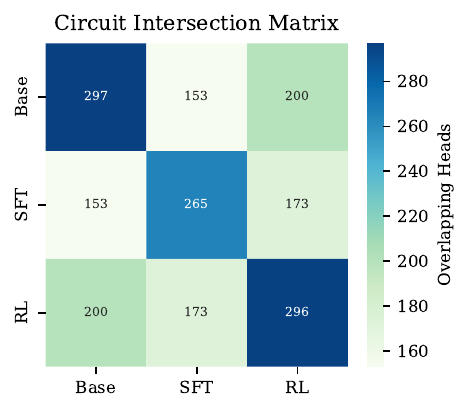}
    \caption{ \textbf{Pairwise Attention Head Overlap Across Base, SFT, and RL Circuits.} Each cell reports the number of attention heads shared between two circuits. Diagonal entries report the total circuit size for each model (Base, SFT, RL). Off-diagonal entries report the number of heads common to the corresponding pair of circuits.
    }
    \label{fig:sft_plot}
\end{figure}

\end{document}